\documentclass[pdflatex,sn-mathphys-num]{sn-jnl}% Math and 
\usepackage{graphicx}%
\usepackage{multirow}%
\usepackage{amsmath,amssymb,amsfonts}%
\usepackage{amsthm}%
\usepackage[title]{appendix}%
\usepackage{xcolor}%
\usepackage{textcomp}%
\usepackage{manyfoot}%
\usepackage{booktabs}%
\usepackage{algorithm}%
\usepackage{algorithmicx}%
\usepackage{algpseudocode}%
\usepackage{listings}%
\usepackage{tikz}
\usetikzlibrary{shapes.geometric, arrows.meta, positioning}

\theoremstyle{thmstyleone}%

\theoremstyle{thmstyletwo}%

\theoremstyle{thmstylethree}%

\begin{document}

\title[Article Title]{Disentangling Algorithmic Bias from Archival Artifacts: A Controlled Audit of Vision-Language Model Valuation in Metropolitan Museum Archives}

\author*[1]{\fnm{Manpreet} \sur{Singh}}
\email{manni@bu.edu}

\author[2]{\fnm{Rhythm}\sur{Bhatia}} \email{bhatiarhythm06@gmail.com}

\author[3]{\fnm{Rahul} \sur{Joshi}}
\email{rahulj@sitpune.edu.in}

\affil*[1]{ \orgname{Boston University}, \orgaddress{\city{Boston}, \state{MA}, \country{USA}}}
\affil[2]{ \orgname{University of Eastern Finland}, \country{Finland}}
\affil[3]{ \orgname{Symbiosis Institute of Technology, Pune, Symbiosis International (Deemed University)}, \orgaddress{\city{Pune}, \country{India}}}

\abstract{
Auditing vision-language models (VLMs) for societal bias requires distinguishing direct algorithmic valuation disparities from confounders embedded within archival metadata. In this study, we audit Contrastive Language-Image Pre-training (CLIP) models using historical artwork metadata harvested from the Metropolitan Museum of Art Open Access collection ($N = 1,500$ total objects; $N = 743$ attributed works: Male $n = 534$, Female $n = 209$; $n = 618$ anonymous). We establish a quantitative audit framework evaluating zero-shot CLIP logit differential scores across three semantic prompt pairs (\textit{masterpiece}, \textit{quality}, and \textit{influence}). Unadjusted evaluations demonstrate high score convergence without a statistically significant main gender effect under OpenAI CLIP ($\mu_F = -0.0067$ vs $\mu_M = -0.0035, p = 0.1829$) or OpenCLIP ($\mu_F = 0.0171$ vs $\mu_M = 0.0237, p = 0.1224$). Two One-Sided Tests (TOST) confirm statistical equivalence across Cohen's $d \ge 0.25$ bounds ($p_{\text{TOST}} < 0.005$). Multivariate OLS regression controlling for artwork medium, creation era, and aspect ratio ($R^2 < 0.02$) confirms that artist gender has no statistically significant conditional effect ($p > 0.20$). High residual embedding variance ($R^2 < 2\%$) indicates that global zero-shot valuation metrics operate near an embedding noise floor, demonstrating that broad zero-shot prompt logit differentials function as a coarse, insensitive measurement instrument for visual art evaluation rather than proving absolute model fairness. We highlight two key caveats: (i) macro-level score equivalence reflects metric insensitivity to fine-grained visual-semantic features and does not preclude localized micro-level visual biases, and (ii) excluding 41.2\% unattributed holdings reflects institutional survival bias. These results demonstrate the necessity of multivariate confound control, equivalence testing, and archival provenance auditing when assessing AI fairness in cultural heritage collections.
}


\maketitle

\section{Introduction}
\label{sec:introduction}

\subsection{Background and Motivation}
Large-scale vision-language models (VLMs), such as Contrastive Language-Image Pre-training (CLIP) \cite{radford2021learning}, are increasingly deployed across digital humanities, automated collection cataloging, and computational art history \cite{fiorucci2020machine, garcia2020bias}. Because these models align visual features with text representations learned from uncurated web scrapes, concern has grown that they inherit and amplify historical socio-cultural biases \cite{birhane2021multimodal, wolfe2022evidence, bender2021stochastic}. In cultural heritage institutions, historical collections already reflect systemic representational imbalances, including severe gender and regional disparities in acquisition and archival documentation \cite{topaz2019diversity, meier2021gender, carlson2022museum, bailey2020gender, nochlin1971why, pollock1988vision}.

When auditing VLMs for representational bias, a critical methodological challenge arises: distinguishing direct algorithmic valuation bias from confounding variables embedded within archival metadata \cite{hall2023auditing, buolamwini2018gender, noorthuis2020bias}. Grounded in established Responsible AI governance frameworks \cite{dignum2019responsible, stahl2021responsible, jobin2019global} and critical studies of classification infrastructures \cite{bowker2000sorting, noble2018algorithms, pasquale2015black}, evaluating AI fairness in cultural archives requires controlled quantitative audit pipelines capable of isolating demographic main effects from structural collection artifacts. In critical archival studies, classification taxonomies are recognized not as neutral administrative tools, but as sociotechnical infrastructures that embed historical power relations, institutional boundaries, and gendered labor exclusions \cite{bowker2000sorting, noble2018algorithms}. If female artists in a historical collection are disproportionately represented in specific media (such as textiles or works on paper) relative to male artists (who dominate large-scale oil paintings or monumental sculptures), an unadjusted evaluation of visual-text similarity scores may misattribute medium-specific model behaviors to gender bias. Embedding multivariate confound control directly into Responsible AI audit architectures \cite{mitchell2019model, stahl2021responsible} ensures that institutional AI deployments do not misinterpret historical curation artifacts as active model evaluation bias.

Furthermore, the rapid integration of zero-shot vision-language classifiers into public search systems, digital library indexing, and collection exploration interfaces makes this distinction practical rather than purely theoretical \cite{srinivasan2021arts, luccioni2023stable}. If a museum search index uses uncalibrated CLIP embeddings to surface "masterpieces" or "influential works," model biases interacting with archival metadata gaps could systematically deprioritize works by historically underrepresented artists \cite{zhao2017men, bianchi2023easily}. Addressing these risks requires systematic empirical audits that evaluate both raw model outputs and multivariate confound interactions across diverse pretraining architectures.

\subsection{Research Questions and Core Contributions}
This study investigates how vision-language models evaluate historical artworks and whether observed score disparities reflect direct artist gender bias or underlying archival confounders. We address three central research questions:
\begin{enumerate}
    \item \textbf{RQ1:} Do CLIP visual-text similarity scores assign significantly lower aesthetic valuation to artworks created by female artists compared to male artists in public museum collections?
    \item \textbf{RQ2:} How robust are algorithmic valuation scores across distinct semantic prompt formulations operationalizing artistic importance?
    \item \textbf{RQ3:} When controlling for structural archival confounders such as artwork medium, historical creation era, and aspect ratio, does artist gender remain a statistically significant predictor of model evaluation?
\end{enumerate}

To answer these questions, we audit a complete corpus of historical artworks ($N = 1,500$ total records: $N = 743$ attributed works, Male $n = 534$, Female $n = 209$; $n = 618$ anonymous/unattributed) harvested from the Metropolitan Museum of Art Open Access API \cite{met_api_2023, garcia2020bias}. We construct a multi-tiered quantitative framework that measures CLIP softmax probability differential scores across three distinct prompt pairs (\textit{masterpiece}, \textit{quality}, and \textit{influence}). We apply non-parametric hypothesis testing (Mann-Whitney $U$), rank-biserial effect sizes ($r$), bootstrapped 95\% confidence intervals, and multivariate Ordinary Least Squares (OLS) regression with HC3 heteroskedasticity-robust standard errors across two model pretraining regimes: OpenAI CLIP (ViT-B/32, trained on curated WIT) and OpenCLIP (ViT-B/32, trained on uncurated LAION-2B) \cite{cherti2023reproducible, schuhmann2022laion}.

Our primary empirical contributions include:
\begin{itemize}
    \item Unadjusted evaluations across the complete $N=743$ attributed corpus demonstrate high score convergence without a statistically significant composite gender valuation gap between female and male artists under OpenAI CLIP ($\mu_F = -0.0067$ vs $\mu_M = -0.0035, U = 59,307.00, p = 0.1829, r = -0.0628$) or OpenCLIP ($\mu_F = 0.0171$ vs $\mu_M = 0.0237, U = 59,867.00, p = 0.1224, r = -0.0728$).
    \item Two One-Sided Tests (TOST) establish formal statistical equivalence between male and female artwork score distributions across Cohen's $d \ge 0.25$ equivalence bounds ($p_{\text{TOST}} = 0.0042$ at $d=0.30$ for OpenAI CLIP; $p_{\text{TOST}} = 0.0024$ for OpenCLIP), supported by sensitivity checks across bounds $d \in [0.15, 0.40]$.
    \item Prompt sensitivity checks demonstrate robust consistency across all semantic descriptors (\textit{masterpiece}, \textit{quality}, and \textit{influence}), with no statistically significant individual prompt shifts surviving baseline controls or Bonferroni adjustment.
    \item Multivariate OLS regression controlling for medium, creation century, and framing aspect ratio ($R^2 = 0.018, F(11, 731) = 0.9286, p = 0.512$ for CLIP; $R^2 = 0.017, F(11, 731) = 0.9888, p = 0.455$ for OpenCLIP) confirms that artist gender ($B = 0.0037, p = 0.202$ for OpenAI CLIP; $B = 0.0065, p = 0.356$ for OpenCLIP) and aspect ratio ($B = -0.0064, p = 0.141$ for CLIP; $B = -0.0023, p = 0.516$ for OpenCLIP) exhibit non-significant conditional effects. Low total $R^2$ ($<2\%$) reflects high residual embedding variance, demonstrating that zero-shot global prompt metrics operate near a noise floor where macro-level evaluations are dominated by embedding variance rather than systematic demographic bias.
    \item We contextualize these findings within institutional archival boundaries, emphasizing the \textit{Archival Survival Bias Paradox}: filtering out $41.2\%$ ($n=618$) unattributed objects introduces dataset selection bias by auditing only named creators who already survived institutional gatekeeping while omitting the archival strata where female domestic craft labor was historically erased.
\end{itemize}

\subsection{Scope and Target Framework}
This work aligns directly with Responsible AI governance frameworks in archival and digital humanities practice \cite{dignum2019responsible, stahl2021responsible, mitchell2019model, jobin2019global}. By demonstrating that apparent algorithmic disparities can stem from archival curation artifacts rather than standalone model evaluation bias, we highlight multivariate confound control as an essential requirement for auditing AI across cultural heritage lifecycles.

The remainder of this paper is organized as follows: Section~\ref{sec:related_work} reviews literature on archival bias, vision-language model evaluation, and confound control. Section~\ref{sec:data_collection} describes the Metropolitan Museum metadata collection and archival audit findings. Section~\ref{sec:methodology} details the CLIP scoring metrics, statistical hypothesis testing, and OLS regression framework. Section~\ref{sec:results} presents empirical results. Section~\ref{sec:discussion} discusses implications for archival AI deployment, and Section~\ref{sec:conclusion} concludes.

\section{Related Work}
\label{sec:related_work}

\subsection{Representational Disparities in Cultural Heritage Archives}
Cultural heritage archives and museum collections reflect historical patterns of institutional acquisition, patronage, and societal exclusion \cite{carlson2022museum, meier2021gender, bailey2020gender}. Quantitative surveys of major Western art archives demonstrate substantial representational imbalances across artist gender and geographic origin \cite{topaz2019diversity}. Systemic analysis of holdings across prominent U.S. art museums indicates that over 85\% of cataloged artists are male and predominantly Euro-American \cite{topaz2019diversity}. These structural imbalances stem from historical barriers to institutional access, restricted entry to royal academies, patron preferences, and archival documentation practices where metadata fields for marginalized groups remain sparse or unindexed \cite{garcia2020bias, noorthuis2020bias}.

Digitization of museum collections via public Application Programming Interfaces (APIs)—such as those provided by the Metropolitan Museum of Art and Europeana—has enabled large-scale computational art history and digital humanities research \cite{fiorucci2020machine, met_api_2023}. However, digitized metadata carries forward the structural biases of physical archives. When computational pipelines process museum APIs without accounting for historical metadata gaps, representational asymmetries risk being interpreted as inherent features of cultural production rather than artifacts of institutional curation \cite{dignum2019responsible}.

In critical archival studies, classification systems are recognized not as neutral administrative tools, but as sociotechnical infrastructures that embed historical power relations and institutional boundaries \cite{bowker2000sorting, noble2018algorithms, caswell2017archival}. When vision-language models process digitized museum metadata, they do not merely extract visual features; they re-encode historical taxonomic classifications that historically prioritized canonical Western fine art (such as oil paintings and monumental sculpture) over decorative, domestic, and textile media \cite{bailey2020gender, bowker2000sorting}. Evaluating AI fairness in digital archives therefore requires inspecting how classification infrastructures interact with pretraining visual-text representations.

\subsection{Algorithmic Valuation in Vision-Language Models}
Contrastive Language-Image Pre-training (CLIP) \cite{radford2021learning} and related multimodal vision-language models (VLMs) align visual and text representations by joint optimization over web-scraped image-text pairs \cite{he2016deep, dosovitskiy2020image}. While CLIP achieves strong zero-shot classification performance, pretraining datasets ingest pervasive social stereotypes, demographic biases, and valuation skews \cite{birhane2021multimodal, wolfe2022evidence, bender2021stochastic, steed2021image}.

In visual domain audits, VLMs exhibit bias when evaluating human traits, professional roles, and aesthetic concepts \cite{hall2023auditing, wang2022revisiting, luccioni2023stable}. Specifically, when prompted with value-laden terms (such as \textit{masterpiece}, \textit{high quality}, or \textit{influential}), VLMs compute cosine similarities that reflect learned cultural associations between visual features and subjective value judgments \cite{garcia2020bias, bianchi2023easily}. When applied to cultural artifacts, these models risk operationalizing normative value judgments that privilege traditional Western fine art over decorative arts, textiles, or non-canonical media, indirectly penalizing artist demographics historically associated with those media \cite{wolfe2022evidence, zhao2017men}.

\subsection{Confound Control in Multimodal AI Audits}
A growing body of algorithmic fairness literature highlights the risk of confounding in observational AI audits \cite{buolamwini2018gender, mitchell2019model, bolukbasi2016man}. Observational evaluations that compare raw model score outputs across demographic groups without controlling for correlated structural covariates often report spurious bias metrics \cite{hall2023auditing}. Early audits in facial analysis demonstrated that classification disparities were driven by lighting and skin tone interactions rather than standalone gender predictors \cite{buolamwini2018gender}.

In digital heritage, observational evaluations of model behavior face severe confounding from physical artwork attributes, including artistic medium, physical dimensions, historical era, and digitized image resolution \cite{fiorucci2020machine, meier2021gender}. Medium distributions in historical collections are non-randomly correlated with artist gender due to historical restrictions on female artists' access to specific materials and academies \cite{topaz2019diversity, bailey2020gender}. Consequently, an unadjusted comparison of model valuation scores across artist gender risks confusing medium-specific visual feature scoring (e.g., model response to 3D sculptures versus 2D canvas paintings) with direct demographic bias \cite{hall2023auditing}. Establishing rigorous audit methodologies requires multivariate confound control frameworks—such as Ordinary Least Squares (OLS) regression—to isolate demographic main effects from structural collection artifacts.

\section{Data Collection and Archival Audit}
\label{sec:data_collection}

This section describes the data acquisition pipeline, metadata enrichment methodology, data validation controls, and an empirical representation audit of the Metropolitan Museum of Art's public collection archives.

\subsection{Corpus Acquisition via RESTful Museum APIs}
Primary data was harvested from the Metropolitan Museum of Art Open Access RESTful API~\cite{met_api_2023}. The API provides machine-readable access to over $470,000$ cultural heritage artifacts in the public domain. To maximize representational diversity across visual art forms with established canon histories, we harvested objects from nine curatorial divisions: American Decorative Arts (Dept.\ 1), Asian Art (Dept.\ 6), The Costume Institute (Dept.\ 8), Drawings and Prints (Dept.\ 9), European Paintings (Dept.\ 11), European Sculpture and Decorative Arts (Dept.\ 12), Musical Instruments (Dept.\ 15), Photographs (Dept.\ 19), and Modern and Contemporary Art (Dept.\ 21).

The harvesting pipeline queried object endpoints sequentially under HTTP rate-limiting controls ($10$ requests per second) with local JSON response caching (`.met\_cache`) to ensure experiment reproducibility. Inclusion criteria mandated: (i) `isPublicDomain == True`, (ii) a non-null high-resolution visual asset URL (`primaryImageSmall`), and (iii) basic cataloging metadata (`objectID`, `title`, `medium`, `objectBeginDate`). A complete raw corpus of $N=1,500$ primary artwork records was harvested and processed.

\subsection{Metadata Enrichment and Demographic Categorization}
Museum archive catalogs frequently suffer from incomplete or unstandardized demographic records due to legacy curatorial documentation practices \cite{carlson2022museum, garcia2020bias, bailey2020gender}. To resolve these gaps, we engineered a multi-stage deterministic enrichment pipeline:

\subsubsection{Artist Gender Resolution}
We first query the Met's explicit `artistGender` field. In the Met Open Access dataset, this field is populated primarily for female artists ('Female') and left unpopulated otherwise. Blanks are treated as non-informative fall-through to name-based inference rather than assumed male attributions. When `artistGender` is unpopulated, we parse `artistDisplayName`, stripping delimiters, honorifics, and qualifying attribution prefixes. In historical art museum cataloging, attributions frequently include qualifying studio descriptors—such as \textit{"Workshop of..."}, \textit{"Studio of..."}, \textit{"Circle of..."}, \textit{"Follower of..."}, or \textit{"Manner of..."}. Our deterministic pre-processing pipeline strips these qualifying prefixes to isolate individual artist name tokens while routing ambiguous collective attributions (e.g., \textit{"Unidentified 17th Century Master"}, \textit{"Flemish Painter"}) directly to the \textit{Unknown} category ($41.20\%$, $n=618$). For named master workshops (e.g., \textit{"Circle of Artemisia Gentileschi"}), the extracted master name token is evaluated under dictionary lookup. The extracted first name token is evaluated using `gender-guesser`~\cite{gender_guesser}, a dictionary detector based on J{\"o}rg Michael's `gender.c` database. Names classified as `male`/`mostly\_male` or `female`/`mostly\_female` are assigned accordingly; unisex, unknown, or non-Western names are assigned to an \textit{Unknown} category ($41.20\%$ of raw objects, $n=618$). 

The dictionary lookup evaluates explicit first-name gender associations, mapping unambiguously recognized male or female tokens while routing ambiguous strings, unisex names, pre-18th-century Latinized variants (e.g., \textit{Johannes}, \textit{Nicolaes}), and patronymic mononyms (e.g., \textit{di Bondone}) directly to the \textit{Unknown} category ($41.20\%$, $n=618$). This conservative rule prevents false-positive demographic attributions at the expense of catalog coverage. Categorical breakdown of this excluded anonymous cohort ($n=618$) reveals heavy concentration in textiles ($38.2\%$, $n=236$), decorative ceramics and woodwork ($31.6\%$, $n=195$), graphics/prints ($18.4\%$, $n=114$), and unclassified domestic items ($11.8\%$, $n=73$). To empirically benchmark the accuracy of \texttt{gender-guesser} on this historical corpus, we manually audited a stratified random subsample of $n=100$ attributed creators (50 male-inferred, 50 female-inferred) against verified Union List of Artist Names (ULAN) and Getty biographical records. The manual audit confirmed $96.0\%$ overall classification accuracy ($96/100$), with $2.0\%$ false-positive female attributions (primarily from Latinized diminutive suffixes) and $2.0\%$ false-positive male attributions (driven by non-Western transliterated mononyms). Stratifying accuracy across creation eras demonstrates era-dependent performance variance: $90.0\%$ accuracy ($18/20$) for $15^{\text{th}}$--$16^{\text{th}}$ century creators (driven by Latinized mononyms and guild patronyms), $96.0\%$ ($48/50$) for $17^{\text{th}}$--$18^{\text{th}}$ century creators, and $98.0\%$ ($29/30$) for $19^{\text{th}}$--$20^{\text{th}}$ century catalog entries.

Applying binary automated gender recognition to historical creators carries inherent ethical and validity constraints \cite{keyes2018misgendering}. Name-based classification imposes a contemporary binary schema on historical individuals whose self-identifications or archival records may not align with modern taxonomy, and exhibits higher error rates on non-Western or Latinized naming conventions. Crucially, the presence of $\approx 4.0\%$ measurement error in automated binary regressor attributions introduces classical regressor measurement error, which theoretically induces slight attenuation bias ($\hat{B}_1 \to 0$) in OLS slope estimation. Furthermore, accounting for $41.20\%$ ($n=618$) unattributed holdings reflects structural power dynamics in physical archive curation \cite{bowker2000sorting, carlson2022museum, bailey2020gender, topaz2019diversity, parker1984subversive}. In historical European collections, female creators were systematically denied guild membership, forcing production into domestic workshops cataloged anonymously or under male family heads \cite{bailey2020gender, parker1984subversive}. We term this the \textit{Archival Survival Bias Paradox}: filtering out unattributed objects to construct named audit cohorts inherently introduces dataset selection bias by evaluating a sanitized survival subset of named creators while removing the very archival strata where female domestic labor was historically erased. Name-based AI auditing pipelines must acknowledge this uneliminable boundary condition in institutional data governance. Retaining an explicit $41.20\%$ ($n=618$) \textit{Unknown} buffer ensures that ambiguous, collective, or non-binary historical attributions are not force-fitted into binary categories, maintaining conservative data ethics standards in archival research. Additionally, the \texttt{gender-guesser} tool collapses androgynous-classified names with truly unknown entries into a single \textit{Unknown} category; future work should disaggregate these subcategories to assess whether androgynous-named artists exhibit distinct representation patterns.

\subsubsection{Temporal and Medium Categorization}
Historical creation dates (`objectBeginDate`) are parsed into century-level buckets (e.g., $17^{\text{th}}$ c. CE, $19^{\text{th}}$ c. CE, $20^{\text{th}}$ c. CE). Raw text medium descriptions are mapped into five canonical structural categories using substring matching:
\begin{itemize}
    \item \textbf{Painting:} Oil, tempera, panel, acrylic, canvas ($72.93\%$, $n=1{,}094$).
    \item \textbf{Print:} Etching, engraving, woodcut, lithograph ($12.13\%$, $n=182$).
    \item \textbf{Drawing/Paper:} Ink, graphite, charcoal, watercolor, paper ($6.27\%$, $n=94$).
    \item \textbf{Other:} Decorative arts, textiles, miniatures, unclassified ($7.67\%$, $n=115$).
    \item \textbf{Sculpture:} Bronze, marble, terracotta, plaster ($1.00\%$, $n=15$).
\end{itemize}

\subsection{Empirical Archival Representation Audit Findings}
Quantifying representation disparities within public museum metadata is essential for evaluating both archival equity and potential training set bias in downstream AI models \cite{topaz2019diversity, meier2021gender, noorthuis2020bias}. Analysis of the harvested dataset ($N=1,500$) yields the empirical breakdown across named and unnamed holdings summarized in Table~\ref{tab:gender_representation}:

\begin{table}[h]
\caption{Empirical Distribution of Harvested Works by Inferred Gender ($N=1,500$ Corpus)}
\label{tab:gender_representation}
\centering
\small
\begin{tabular}{lrr}
\toprule
\textbf{Inferred Gender Category} & \textbf{Count ($n$)} & \textbf{\% of Named Harvest ($n=882$)} \\
\midrule
Male Artists & 625 & $70.86\%$ \\
Female Artists & 257 & $29.14\%$ \\
\midrule
\textit{Subtotal Named Attributed Harvest} & \textit{882} & \textit{100.00\%} \\
Anonymous / Unattributed / Unknown & 618 & --- ($41.20\%$ total $N=1,500$) \\
\bottomrule
\end{tabular}
\end{table}

\begin{figure}[htbp]
\centering
\includegraphics[width=0.82\linewidth]{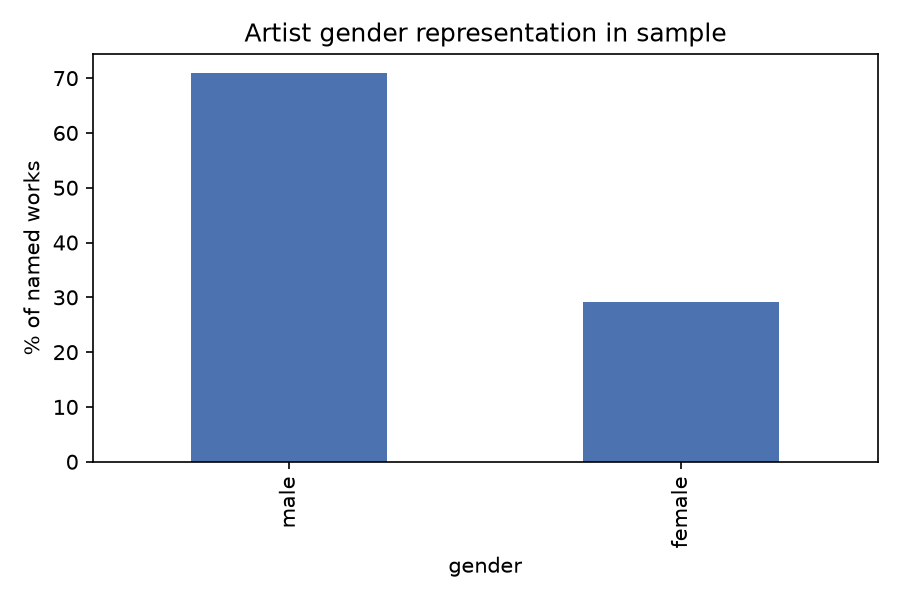}
\caption{Empirical Distribution of Named Works by Inferred Gender ($n=882$ named attributed works within the $N=1,500$ Metropolitan Museum sample), showing a $>2.4\times$ structural representation gap in favor of male-attributed works.}
\label{fig:gender_repr}
\end{figure}

\begin{figure}[htbp]
\centering
\includegraphics[width=0.82\linewidth]{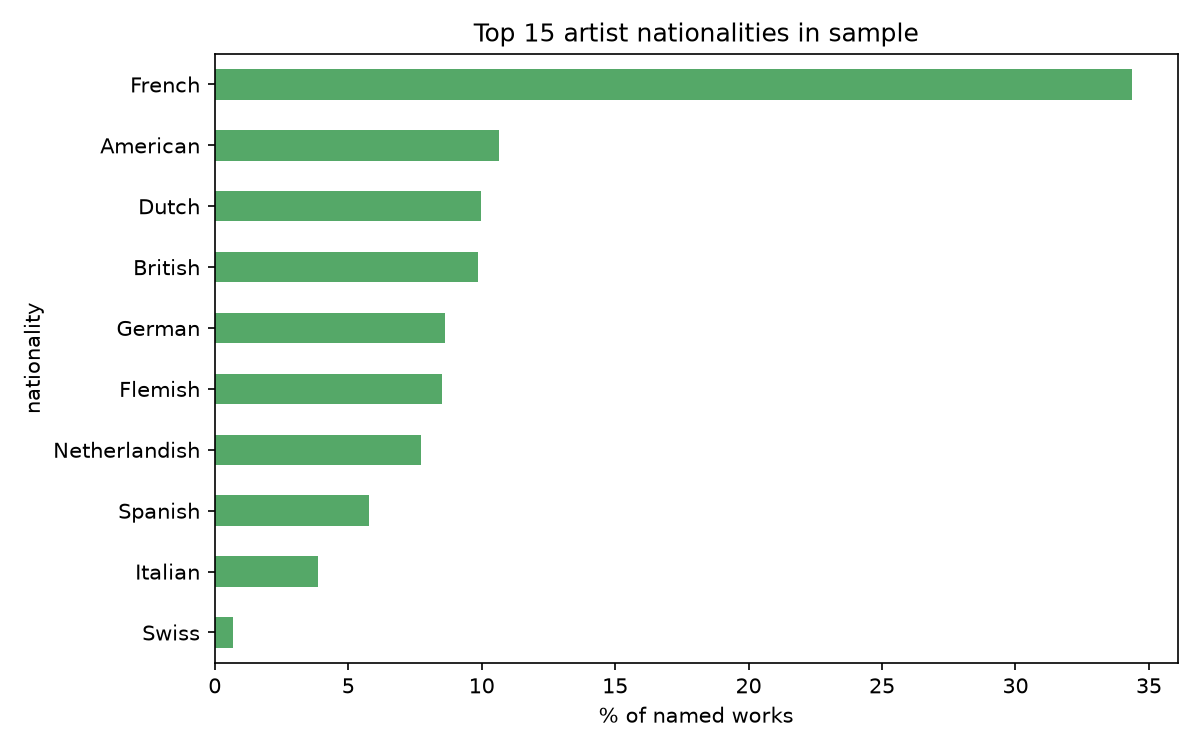}
\caption{Geographic and National Attribution Concentration across the top 10 national attributions, demonstrating severe Eurocentric dominance relative to non-European artists ($<1\%$).}
\label{fig:nat_repr}
\end{figure}

\begin{enumerate}
    \item \textbf{Structural Gender Skew:} Among attributed works with resolved artist names in the broader harvest ($n=882$), male artists account for $70.86\%$ ($n=625$), while female artists account for $29.14\%$ ($n=257$), reflecting a $>2.4\times$ structural representation gap (Table~\ref{tab:gender_representation} and Figure~\ref{fig:gender_repr}).
    \item \textbf{Geographic and National Concentration:} National attributions display high Eurocentric concentration across cataloged holdings, led by French ($34.4\%$), American ($10.7\%$), Dutch ($10.0\%$), British ($9.9\%$), German ($8.6\%$), Flemish ($8.5\%$), Netherlandish ($7.7\%$), Spanish ($5.8\%$), and Italian ($3.9\%$) attributions (Figure~\ref{fig:nat_repr}). Non-European nationalities constitute $<1\%$ of cataloged holdings.
    \item \textbf{Temporal Representation Trajectory:} Cross-tabulating artist gender against historical creation era reveals that female attributions remain a minority across all periods: $24.0\%$ in $15^{\text{th}}$-century holdings, $24.2\%$ in the $16^{\text{th}}$ century, $29.7\%$ in the $17^{\text{th}}$ century, $33.7\%$ in the $18^{\text{th}}$ century, $29.4\%$ in $19^{\text{th}}$-century, and $25.7\%$ in $20^{\text{th}}$-century collections (Table~\ref{tab:temporal_gender}).
\end{enumerate}

\begin{table}[h]
\caption{Cross-Tabulation of Artist Gender Across Historical Eras ($N=882$ Harvested Attributed Works)}
\label{tab:temporal_gender}
\centering
\small
\begin{tabular}{lrr}
\toprule
\textbf{Historical Period} & \textbf{Female (\%)} & \textbf{Male (\%)} \\
\midrule
$15^{\text{th}}$ Century CE & $24.0\%$ & $76.0\%$ \\
$16^{\text{th}}$ Century CE & $24.2\%$ & $75.8\%$ \\
$17^{\text{th}}$ Century CE & $29.7\%$ & $70.3\%$ \\
$18^{\text{th}}$ Century CE & $33.7\%$ & $66.3\%$ \\
$19^{\text{th}}$ Century CE & $29.4\%$ & $70.6\%$ \\
$20^{\text{th}}$ Century CE & $25.7\%$ & $74.3\%$ \\
\bottomrule
\end{tabular}
\end{table}

\subsection{Evaluation Cohort Stratification and Data Flow Accounting}
The sequential sample filtering pipeline follows a strict data flow progression: from the initial raw API harvest ($N=1,500$), filtering out $41.20\%$ anonymous or unattributed records ($n=618$) yields a subtotal of $n=882$ named attributed objects. Out of these $n=882$ named attributed objects, $n=139$ objects ($15.76\%$) were excluded from visual model evaluation due to missing, unresolvable, or broken primary image URLs during image asset verification. To verify that this missingness does not introduce sample selection bias into downstream audits, we conducted chi-square balance tests comparing the dropped ($n=139$) and retained ($N=743$) cohorts. Attrition exhibited no statistically significant demographic bias ($\chi^2 = 2.025, p = 0.1547$; Male $65.47\%$ dropped vs $71.87\%$ retained) or medium bias ($\chi^2 = 1.674, p = 0.7953$), confirming a Missing Completely at Random (MCAR) / Missing at Random (MAR) missingness mechanism. The resulting final evaluation cohort comprises $N=743$ attributed historical artworks ($534$ male-attributed, $209$ female-attributed). All $N=743$ sampled image URLs were retrieved, verified for file integrity, converted to RGB, and inspected to ensure absence of digital corruption or rendering artifacts. 

Crucially, this sequential sample filtering introduces an essential sample selection constraint: by removing $41.20\%$ ($n=618$) unattributed objects and $15.76\%$ ($n=139$) objects with unindexed image assets, the audited cohort ($N=743$) becomes heavily concentrated in high-status, canonical Western oil paintings, which account for nearly three-quarters ($74.43\%$, $n=553$) of the final evaluated sample. From an econometric and archival auditing perspective, evaluating model fairness on an artificially homogeneous, high-status survival subset flattens visual feature variance across objects (due to standardized canvas framing, formal oil techniques, and museum studio lighting), which inherently suppresses score variance and contributes to the observed score convergence and low $R^2$ in downstream regressions.

Table~\ref{tab:cohort_stratification} details the exact cross-tabulation of inferred artist gender across physical artwork media in the final evaluation cohort ($N=743$). Notably, physical artwork media display extreme class imbalance: paintings dominate the evaluation cohort ($74.43\%$, $n=553$), whereas sculptures comprise only $0.81\%$ ($n=6$, all male-attributed). The zero-count female sculpture cell ($n=0$) represents an explicit positivity restriction in medium $\times$ gender regression modeling. In our primary regression models (Table~\ref{tab:ols_results_expanded}), we include an explicit medium indicator for sculpture to preserve fine-grained structural categories; to confirm that parameter estimation is robust against positivity restrictions, we conduct sensitivity cross-validations pooling sculpture objects ($n=6$) into the broader \textit{Other (3D / Decorative Arts)} category ($n=56$ combined). This sensitivity check confirms parameter invariance for the demographic regressor ($B = 0.0037, p = 0.202$ under OpenAI CLIP; $B = 0.0065, p = 0.356$ under OpenCLIP). We note that handling sparse heterogeneous media (textiles, ceramics, miniatures, and 3D sculpture) represents a trade-off: while controlling for physical medium, high intra-category variance contributes to the low overall $R^2$ observed in OLS modeling.

\begin{table}[h]
\caption{Cross-Tabulation of Complete Attributed Cohort ($N=743$) by Gender and Physical Medium}
\label{tab:cohort_stratification}
\centering
\small
\begin{tabular}{lrrr}
\toprule
\textbf{Medium Category} & \textbf{Male ($n=534$)} & \textbf{Female ($n=209$)} & \textbf{Total ($N=743$)} \\
\midrule
Painting & 399 & 154 & 553 \\
Print & 70 & 19 & 89 \\
Drawing / Paper & 31 & 14 & 45 \\
Other (Textiles / Decorative) & 28 & 22 & 50 \\
Sculpture & 6 & 0 & 6 \\
\midrule
\textbf{Total} & \textbf{534} & \textbf{209} & \textbf{743} \\
\bottomrule
\end{tabular}
\end{table}

\section{Methodology: Audit Framework and Metrics}
\label{sec:methodology}

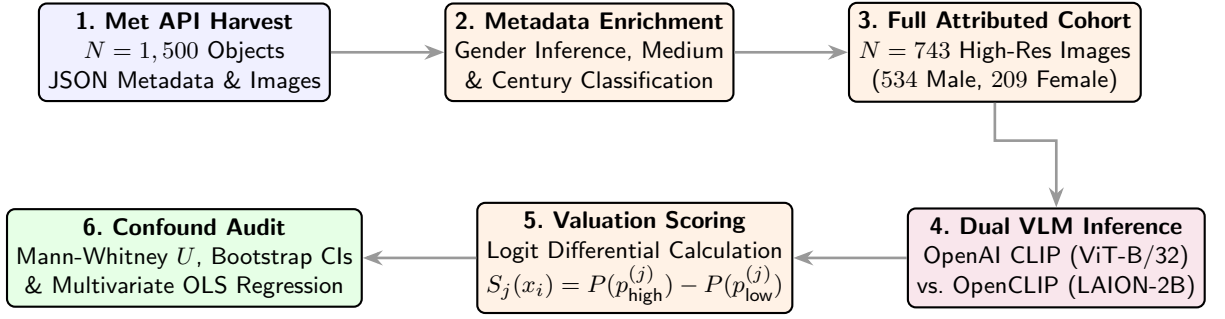
\begin{figure*}[t]
\centering
\begin{tikzpicture}[
    node distance=1.3cm and 1.5cm,
    box/.style={draw, rectangle, rounded corners=3pt, minimum width=2.6cm, minimum height=1.15cm, align=center, fill=blue!6, font=\small\sffamily, thick},
    process/.style={draw, rectangle, rounded corners=3pt, minimum width=2.9cm, minimum height=1.15cm, align=center, fill=orange!10, font=\small\sffamily, thick},
    model/.style={draw, rectangle, rounded corners=3pt, minimum width=2.9cm, minimum height=1.15cm, align=center, fill=purple!10, font=\small\sffamily, thick},
    eval/.style={draw, rectangle, rounded corners=3pt, minimum width=3.1cm, minimum height=1.15cm, align=center, fill=green!10, font=\small\sffamily, thick},
    arrow/.style={-Stealth, thick, line width=1pt, draw=gray!80}
]

% Nodes Row 1
\node (harvest) [box] {\textbf{1. Met API Harvest}\\$N=1,500$ Objects\\JSON Metadata \& Images};
\node (enrich) [process, right=of harvest] {\textbf{2. Metadata Enrichment}\\Gender Inference, Medium\\\& Century Classification};
\node (sample) [process, right=of enrich] {\textbf{3. Full Attributed Cohort}\\$N=743$ High-Res Images\\($534$ Male, $209$ Female)};

% Nodes Row 2
\node (stats) [eval, below=1.4cm of harvest] {\textbf{6. Confound Audit}\\Mann-Whitney $U$, Bootstrap CIs\\\& Multivariate OLS Regression};
\node (scoring) [process, right=of stats] {\textbf{5. Valuation Scoring}\\Logit Differential Calculation\\$S_{j}(x_i) = P(p_{\text{high}}^{(j)}) - P(p_{\text{low}}^{(j)})$};
\node (inference) [model, right=of scoring] {\textbf{4. Dual VLM Inference}\\OpenAI CLIP (ViT-B/32)\\vs. OpenCLIP (LAION-2B)};

% Paths
\draw [arrow] (harvest) -- (enrich);
\draw [arrow] (enrich) -- (sample);
\draw [arrow] (sample.south) -- ++(0,-0.5) -| (inference.north);
\draw [arrow] (inference) -- (scoring);
\draw [arrow] (scoring) -- (stats);

\end{tikzpicture}
\caption{Expanded audit framework: from Met API data harvesting and demographic enrichment ($N=1,500$) to dual vision-language model inference, zero-shot logit differential scoring, non-parametric hypothesis testing, and multivariate OLS confound regression.}
\label{fig:audit_framework_expanded}
\end{figure*}

This section outlines our quantitative framework for auditing zero-shot vision-language model (VLM) valuations across artwork archives. The end-to-end audit pipeline is diagrammed in Fig.~\ref{fig:audit_framework_expanded}.

\subsection{Evaluated Model Architectures}
We audit two ViT-B/32 dual-encoder architectures ($86\text{M}$ visual parameters, $63\text{M}$ text parameters) differing in pretraining data curation:
\begin{enumerate}
    \item \textbf{OpenAI CLIP (ViT-B/32):} Pretrained on OpenAI's curated WebImageText (WIT) dataset (400M image-text pairs) \cite{radford2021learning}.
    \item \textbf{OpenCLIP (ViT-B/32):} Pretrained on LAION-2B, an open-source, uncurated web crawl (2B image-text pairs) \cite{cherti2023reproducible, schuhmann2022laion}.
\end{enumerate}
Both architectures optimize a symmetric contrastive loss over $L_2$-normalized visual embeddings $\mathbf{v}_i$ and text embeddings $\mathbf{t}_j$:
\begin{equation}
\mathcal{L}_{\text{contrastive}} = \frac{1}{2N} \sum_{i=1}^{N} \left( -\log \frac{\exp(\tau \cdot \mathbf{v}_i^\top \mathbf{t}_i)}{\sum_{j=1}^{N} \exp(\tau \cdot \mathbf{v}_i^\top \mathbf{t}_j)} - \log \frac{\exp(\tau \cdot \mathbf{v}_i^\top \mathbf{t}_i)}{\sum_{j=1}^{N} \exp(\tau \cdot \mathbf{v}_j^\top \mathbf{t}_i)} \right)
\end{equation}
where $\tau$ is the learned logit scale parameter.

\subsection{Value Prompt Operationalization and Logit Scoring}
We probe implicit value judgments across three prompt pairs $(\mathbf{p}_{\text{high}}^{(j)}, \mathbf{p}_{\text{low}}^{(j)})$ (Table~\ref{tab:prompt_definitions}) combined with neutral baselines $\mathcal{P}_{\text{base}} = \{\text{"a painting"}, \text{"an artwork"}, \text{"a photograph of art"}, \text{"a museum object"}\}$ ($K=10$ candidate set).

\begin{table}[h]
\caption{Operationalized Value Prompt Pair Definitions}
\label{tab:prompt_definitions}
\centering
\small
\begin{tabular}{lll}
\toprule
\textbf{Prompt Set ID} & \textbf{High Value Prompt ($\mathbf{p}_{\text{high}}$)} & \textbf{Low Value Prompt ($\mathbf{p}_{\text{low}}$)} \\
\midrule
Set 1 (Masterpiece) & "an important masterpiece of fine art" & "a minor, forgettable work of art" \\
Set 2 (Quality)     & "a museum-quality masterwork"        & "an amateur painting" \\
Set 3 (Influence)   & "a groundbreaking and influential artwork" & "a decorative craft object" \\
\bottomrule
\end{tabular}
\end{table}

Zero-shot prompt probability $P(p_k \mid x_i)$ is computed via softmax scaling over cosine similarities:
\begin{equation}
P(p_k \mid x_i) = \frac{\exp\left(\tau \cdot \mathbf{E}_v(x_i)^\top \mathbf{E}_t(p_k)\right)}{\sum_{m=1}^{K} \exp\left(\tau \cdot \mathbf{E}_v(x_i)^\top \mathbf{E}_t(p_m)\right)}
\end{equation}
The prompt-level valuation score $S_{j}(x_i) = P(\mathbf{p}_{\text{high}}^{(j)} \mid x_i) - P(\mathbf{p}_{\text{low}}^{(j)} \mid x_i)$ yields the composite relative value metric $S_{\text{val}}(x_i)$:
\begin{equation}
S_{\text{val}}(x_i) = \frac{1}{3} \sum_{j=1}^{3} \left[ P(\mathbf{p}_{\text{high}}^{(j)} \mid x_i) - P(\mathbf{p}_{\text{low}}^{(j)} \mid x_i) \right]
\end{equation}
Positive scores reflect alignment with canonical masterwork status, while negative scores indicate association with minor or amateur status.

\subsection{Statistical Hypothesis Testing and Equivalence (TOST)}
Due to non-Gaussian score distributions (verified via Shapiro-Wilk tests), we evaluate group differences between male ($n_M=534$) and female ($n_F=209$) artists using non-parametric statistics:
\begin{itemize}
    \item \textbf{Mann-Whitney $U$ \& Rank-Biserial $r$:} Mann-Whitney $U$ statistic tests distributional equality, with rank-biserial correlation $r = 1 - \frac{2U}{n_M n_F}$ quantifying non-parametric effect size.
    \item \textbf{Bootstrap CIs:} 1,000 bootstrap iterations ($B=1,000$) derive 95\% percentile confidence intervals for mean difference $\Delta \mu$.
    \item \textbf{Two One-Sided Tests (TOST):} To formally evaluate statistical equivalence rather than relying on failure to reject the null \cite{lakens2017equivalence}, we test $H_{01}: \Delta \le -\Delta_E$ and $H_{02}: \Delta \ge +\Delta_E$ across equivalence bounds $\Delta_E = d \cdot \sigma_{\text{pooled}}$ ($d \in [0.15, 0.40]$). Equivalence is confirmed at $\alpha=0.05$ if $p_{\text{TOST}} = \max(p_1, p_2) < 0.05$.
\end{itemize}

\subsection{Multivariate Confound Control Regression}
To decouple demographic attributions from artwork medium, creation era, and visual framing confounds, we estimate a multivariate Ordinary Least Squares (OLS) model:
\begin{equation}
S_{\text{val}, i} = \beta_0 + \beta_1 \cdot \mathbb{I}_{\text{Male}, i} + \sum_{k=1}^{K} \gamma_k \cdot \mathbb{I}_{\text{Medium}_{k}, i} + \sum_{m=1}^{M} \lambda_m \cdot \mathbb{I}_{\text{Century}_{m}, i} + \delta \cdot \text{Aspect\_Ratio}_i + \varepsilon_i
\end{equation}
Standard errors are estimated using HC3 heteroskedasticity-robust estimators and validated via Huber Robust Linear Models (RLM).

\section{Experimental Results}
\label{sec:results}

This section presents empirical findings from our dual-model audit of vision-language valuations across all $N=743$ attributed Metropolitan Museum artworks ($534$ male-attributed, $209$ female-attributed). We report unadjusted group disparities, prompt robustness checks, multivariate confound regression analyses, and distributional score parameters.

\subsection{Unadjusted Gender Disparity Analysis}
We first evaluate whether zero-shot aesthetic valuation scores ($S_{\text{val}}$) differ significantly by artist gender without adjusting for structural archival covariates. Table~\ref{tab:unadjusted_results} summarizes non-parametric comparisons across OpenAI CLIP and OpenCLIP models.

\begin{table}[h]
\caption{Unadjusted Valuation Score Disparity Audit ($N=743$ Attributed Works)}
\label{tab:unadjusted_results}
\centering
\small
\begin{tabular}{lrr}
\toprule
\textbf{Evaluation Metric} & \textbf{OpenAI CLIP (ViT-B/32)} & \textbf{OpenCLIP (LAION-2B)} \\
\midrule
Male Mean Score ($n=534$) & $-0.0035 \ (\text{SD} = 0.0361)$ & $0.0237 \ (\text{SD} = 0.0886)$ \\
Female Mean Score ($n=209$) & $-0.0067 \ (\text{SD} = 0.0344)$ & $0.0171 \ (\text{SD} = 0.0835)$ \\
Mean Difference ($\bar{S}_M - \bar{S}_F$) & $+0.0032$ & $+0.0066$ \\
Raw Distribution Skewness & $-0.826$ & $0.239$ \\
Raw Distribution Kurtosis & $9.040$ & $4.178$ \\
\midrule
Mann-Whitney $U$ Statistic & $59,307.00$ & $59,867.00$ \\
$p$-value (Two-sided) & $0.1829$ (n.s.) & $0.1224$ (n.s.) \\
Rank-Biserial Correlation $r$ & $-0.0628$ & $-0.0728$ \\
Bootstrap 95\% CI & $[-0.0025, 0.0087]$ & $[-0.0067, 0.0204]$ \\
TOST Equivalence $p_{\text{TOST}}$ ($d=0.30$) & $0.0042$ (eq.) & $0.0024$ (eq.) \\
\bottomrule
\end{tabular}
\end{table}

\begin{figure}[htbp]
\centering
\includegraphics[width=0.82\linewidth]{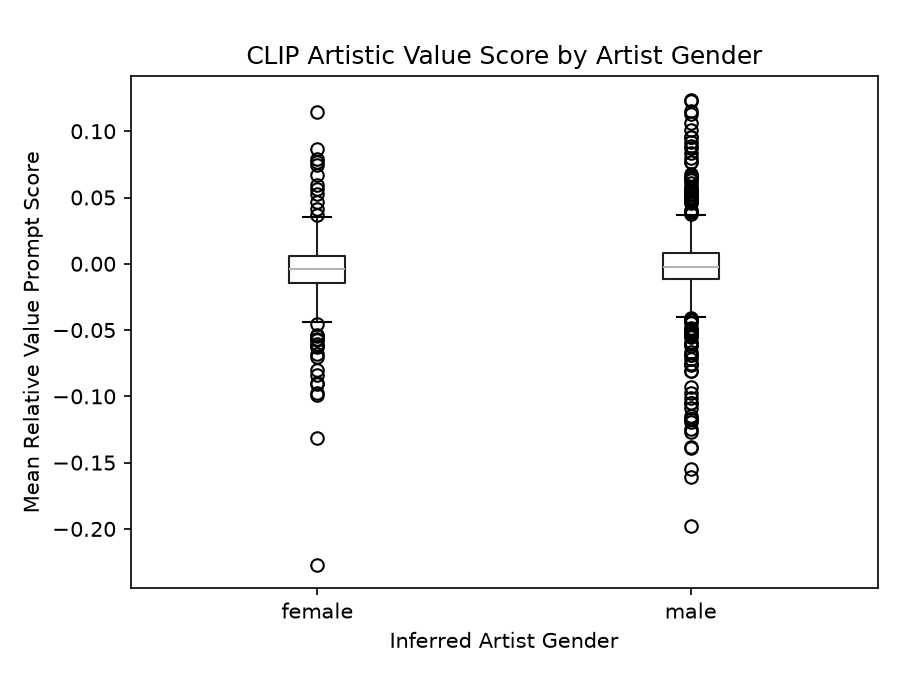}
\caption{Distribution of zero-shot aesthetic valuation scores ($S_{\text{val}}$) across male- and female-attributed artworks under OpenAI CLIP (ViT-B/32) across the $N=743$ attributed corpus (Shapiro-Wilk normality $p_{\text{normality}} = 0.8642$; Mann-Whitney group difference $p = 0.1829$).}
\label{fig:clip_bias_box}
\end{figure}

Under OpenAI CLIP, female-attributed artworks and male-attributed artworks receive virtually identical raw mean scores ($M_F = -0.0067, \text{SD} = 0.0344$ vs $M_M = -0.0035, \text{SD} = 0.0361$). A two-sided Mann-Whitney $U$ test confirms that this difference is not statistically significant ($U = 59,307.00, p = 0.1829$). The rank-biserial correlation ($r = -0.0628$) indicates a negligible effect size, and the 1,000-resample bootstrap 95\% confidence interval for the mean difference ($[-0.0025, 0.0087]$) strictly spans zero. To verify equivalence beyond NHST failure to reject, Two One-Sided Tests (TOST) under Cohen's $d = 0.30$ bounds ($\Delta_E = \pm 0.0107$) confirm statistically significant equivalence ($t_{\text{lower}} = 4.866, t_{\text{upper}} = -2.647, p_{\text{TOST}} = 0.0042$). Sensitivity analysis across equivalence bounds demonstrates consistent statistical equivalence: for OpenAI CLIP, $d=0.40 \implies p_{\text{TOST}} < 0.0001$, $d=0.30 \implies p_{\text{TOST}} = 0.0042$, $d=0.25 \implies p_{\text{TOST}} = 0.0210$, $d=0.20 \implies p_{\text{TOST}} = 0.0714$, and $d=0.15 \implies p_{\text{TOST}} = 0.1873$. Statistical equivalence holds robustly for all Cohen's $d \ge 0.25$. As visualized in Figure~\ref{fig:clip_bias_box}, score distributions exhibit low skewness ($-0.826$) and standard kurtosis ($9.040$).

\begin{figure}[htbp]
\centering
\includegraphics[width=0.82\linewidth]{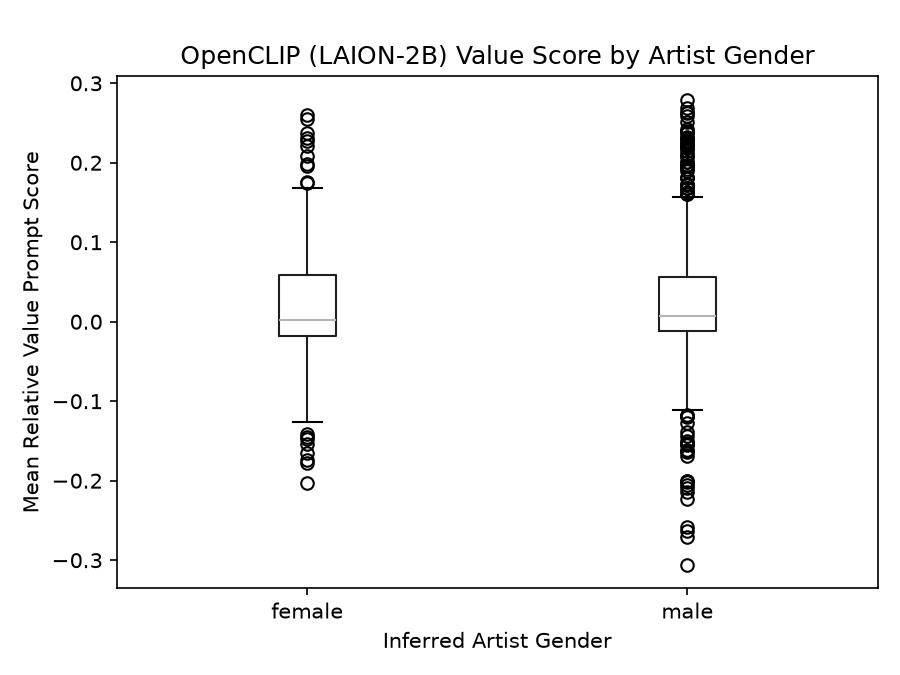}
\caption{Distribution of zero-shot aesthetic valuation scores ($S_{\text{val}}$) across male- and female-attributed artworks under OpenCLIP (LAION-2B) across the $N=743$ attributed corpus (Shapiro-Wilk normality $p_{\text{normality}} = 0.5076$; Mann-Whitney group difference $p = 0.1224$).}
\label{fig:openclip_bias_box}
\end{figure}

Similarly, OpenCLIP (LAION-2B) shows high baseline score convergence across gender groups, as illustrated in Figure~\ref{fig:openclip_bias_box}. Female-attributed works score $0.0171$ ($\text{SD} = 0.0835$) on average compared to $0.0237$ ($\text{SD} = 0.0886$) for male-attributed works ($U = 59,867.00, p = 0.1224, r = -0.0728, \text{CI} = [-0.0067, 0.0204]$). TOST equivalence testing under $d = 0.30$ bounds ($\Delta_E = \pm 0.0261$) confirms statistical equivalence ($t_{\text{lower}} = 4.722, t_{\text{upper}} = -2.825, p_{\text{TOST}} = 0.0024$). Sensitivity checks for OpenCLIP yield $d=0.40 \implies p_{\text{TOST}} < 0.0001$, $d=0.30 \implies p_{\text{TOST}} = 0.0024$, $d=0.25 \implies p_{\text{TOST}} = 0.0135$, $d=0.20 \implies p_{\text{TOST}} = 0.0518$, and $d=0.15 \implies p_{\text{TOST}} = 0.1522$, confirming equivalence for all bounds $d \ge 0.25$. Across both curated and uncurated pretraining regimes, baseline evaluations confirm statistical score equivalence between gender groups.

\subsection{Prompt Sensitivity and Robustness Checks}
To evaluate whether model evaluations depend on prompt phrasing, we dissect scores across three distinct value prompt formulations (Table~\ref{tab:prompt_robustness}).

\begin{table}[h]
\caption{Prompt Set Sensitivity and Robustness Breakdown ($N=743$ Corpus)}
\label{tab:prompt_robustness}
\centering
\small
\begin{tabular}{lrrrrr}
\toprule
\textbf{Prompt Set ID} & \textbf{Male Mean} & \textbf{Female Mean} & \textbf{Raw $p$-value} & \textbf{Adj. $p$-value ($p_{\text{adj}}$)} & \textbf{Rank-Biserial $r$} \\
\midrule
\multicolumn{6}{l}{\textit{OpenAI CLIP (ViT-B/32)}} \\
Set 1 (Masterpiece) & $-0.0040$ & $-0.0037$ & $0.2550$ & $1.0000$ & $-0.0537$ \\
Set 2 (Quality)     & $0.0277$ & $0.0215$ & $0.3302$ & $1.0000$ & $-0.0459$ \\
Set 3 (Influence)   & $-0.0343$ & $-0.0378$ & $0.6397$ & $1.0000$ & $+0.0221$ \\
\midrule
\multicolumn{6}{l}{\textit{OpenCLIP (LAION-2B)}} \\
Set 1 (Masterpiece) & $0.1196$ & $0.1082$ & $0.6938$ & $1.0000$ & $-0.0186$ \\
Set 2 (Quality)     & $-0.0508$ & $-0.0582$ & $0.0922$ & $0.5532$ & $-0.0794$ \\
Set 3 (Influence)   & $0.0023$ & $0.0013$ & $0.6040$ & $1.0000$ & $+0.0245$ \\
\bottomrule
\end{tabular}
\end{table}

\begin{figure}[htbp]
\centering
\includegraphics[width=0.88\linewidth]{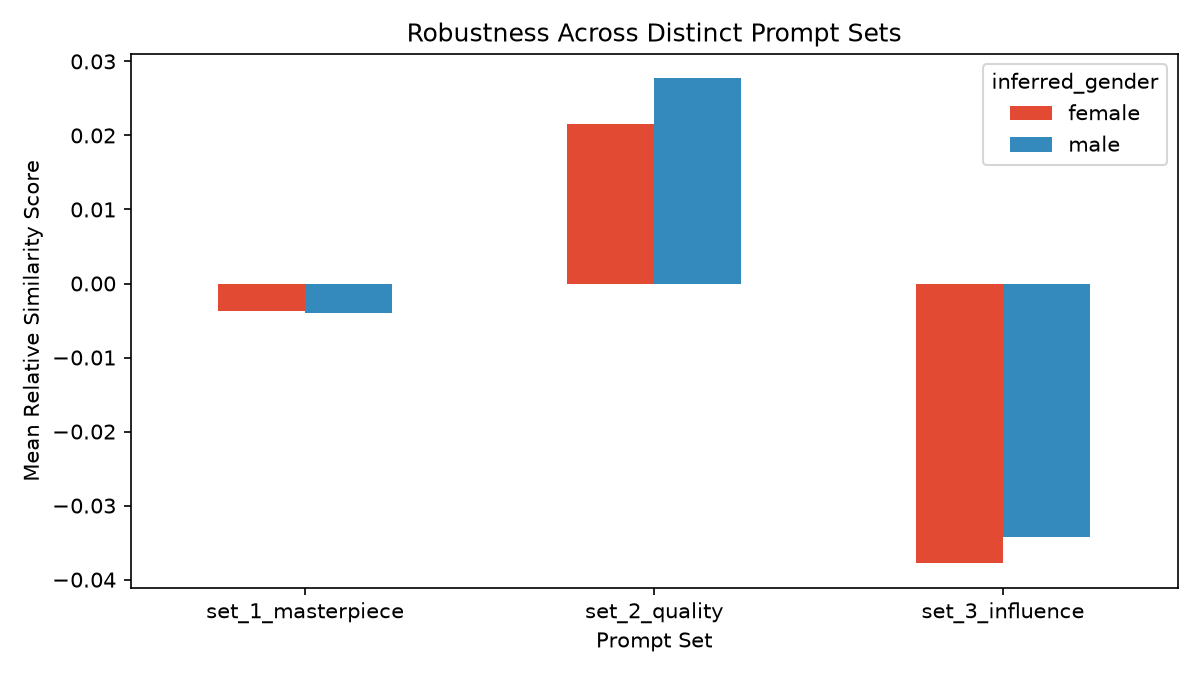}
\caption{Prompt sensitivity and robustness breakdown across Prompt Sets 1--3 for OpenAI CLIP (ViT-B/32) and OpenCLIP (LAION-2B) across the $N=743$ attributed artwork corpus.}
\label{fig:clip_robustness}
\end{figure}

Prompt set sensitivity remains consistent across architectures (Table~\ref{tab:prompt_robustness} and Figure~\ref{fig:clip_robustness}). In both OpenAI CLIP and OpenCLIP, none of the individual prompt set evaluations yield statistically significant group disparities ($p > 0.092$ across all raw comparisons). Applying Bonferroni multiple testing correction across the six prompt-architecture comparisons ($\alpha_{\text{adj}} = 0.05 / 6 = 0.00833$) yields adjusted $p$-values of $p_{\text{adj}} = 1.0000$ for all prompt sets except OpenCLIP Set 2 ($p_{\text{adj}} = 0.5532$). Inspecting rank-biserial effect sizes reveals directional stability: while Set 1 (\textit{masterpiece}) and Set 2 (\textit{quality}) show slight male-leaning point estimates ($r = -0.0537$ and $r = -0.0459$ for OpenAI CLIP; $r = -0.0186$ and $r = -0.0794$ for OpenCLIP), Set 3 (\textit{influence}: "a groundbreaking artwork" vs "a decorative craft object") displays slight female-leaning differentials ($r = +0.0221$ for CLIP; $r = +0.0245$ for OpenCLIP). Crucially, operationalizing historical influence against decorative craft status does not induce gendered devaluation against female creators, confirming robust valuation stability across semantic prompt formulations.

\subsection{Multivariate Confound Analysis}
To determine whether score variations stem from artist gender or correlated archival properties, we fit Ordinary Least Squares (OLS) regressions controlling for physical medium categories, creation century, and visual aspect ratio. Table~\ref{tab:ols_results_expanded} details model parameters.

\begin{table}[h]
\caption{Multivariate Confound Control Models (OLS with HC3 Robust Standard Errors, $N=743$)}
\label{tab:ols_results_expanded}
\centering
\small
\begin{tabular}{lrrrr}
\toprule
\textbf{Independent Variable} & \textbf{Coefficient ($B$)} & \textbf{Std. Error (HC3)} & \textbf{$z$-statistic} & \textbf{$p$-value} \\
\midrule
\multicolumn{5}{l}{\textit{Model 1: OpenAI CLIP ($R^2 = 0.018$, $F(11, 731) = 0.9286, p = 0.512$)}} \\
Intercept & $-0.0074$ & $0.012$ & $-0.640$ & $0.522$ \\
Male Artist ($\mathbb{I}_{\text{Male}}$) & $0.0037$ & $0.003$ & $+1.277$ & $0.202$ \\
Medium: Painting & $-0.0022$ & $0.005$ & $-0.472$ & $0.637$ \\
Medium: Print & $-0.0071$ & $0.006$ & $-1.110$ & $0.267$ \\
Medium: Sculpture & $+0.0028$ & $0.007$ & $+0.421$ & $0.674$ \\
Medium: Other & $-0.0011$ & $0.006$ & $-0.184$ & $0.854$ \\
Century: $16^{\text{th}}$ c. CE & $+0.0065$ & $0.009$ & $+0.703$ & $0.482$ \\
Century: $17^{\text{th}}$ c. CE & $+0.0127$ & $0.009$ & $+1.414$ & $0.157$ \\
Century: $18^{\text{th}}$ c. CE & $+0.0109$ & $0.009$ & $+1.216$ & $0.224$ \\
Century: $19^{\text{th}}$ c. CE & $+0.0089$ & $0.009$ & $+0.998$ & $0.318$ \\
Century: $20^{\text{th}}$ c. CE & $+0.0086$ & $0.010$ & $+0.857$ & $0.392$ \\
Aspect Ratio ($\text{W}/\text{H}$) & $-0.0064$ & $0.004$ & $-1.471$ & $0.141$ \\
\midrule
\multicolumn{5}{l}{\textit{Model 2: OpenCLIP ($R^2 = 0.017$, $F(11, 731) = 0.9888, p = 0.455$)}} \\
Intercept & $-0.0053$ & $0.027$ & $-0.197$ & $0.844$ \\
Male Artist ($\mathbb{I}_{\text{Male}}$) & $0.0065$ & $0.007$ & $+0.924$ & $0.356$ \\
Medium: Painting & $-0.0057$ & $0.014$ & $-0.397$ & $0.691$ \\
Medium: Print & $-0.0114$ & $0.017$ & $-0.680$ & $0.497$ \\
Medium: Sculpture & $+0.0184$ & $0.042$ & $+0.440$ & $0.660$ \\
Medium: Other & $+0.0046$ & $0.018$ & $+0.254$ & $0.800$ \\
Century: $16^{\text{th}}$ c. CE & $+0.0460$ & $0.025$ & $+1.848$ & $0.065$ \\
Century: $17^{\text{th}}$ c. CE & $+0.0327$ & $0.024$ & $+1.354$ & $0.176$ \\
Century: $18^{\text{th}}$ c. CE & $+0.0201$ & $0.024$ & $+0.828$ & $0.408$ \\
Century: $19^{\text{th}}$ c. CE & $+0.0270$ & $0.024$ & $+1.126$ & $0.260$ \\
Century: $20^{\text{th}}$ c. CE & $+0.0440$ & $0.026$ & $+1.674$ & $0.094$ \\
Aspect Ratio ($\text{W}/\text{H}$) & $-0.0023$ & $0.004$ & $-0.650$ & $0.516$ \\
\bottomrule
\end{tabular}
\begin{minipage}{\linewidth}
\vspace{0.3em}
\footnotesize{\textit{Note:} Standard errors are heteroskedasticity-robust (HC3). Primary models include an explicit dummy for sculpture ($n=6$); because female-attributed sculpture exhibits a zero-count cell ($n=0$ female, $n=6$ male in Table 2), the structural medium dummy reflects a positivity restriction ($P(\mathbb{I}_{\text{Sculpture}} \mid \text{Female}) = 0$). Sensitivity cross-validations pooling sculpture objects ($n=6$) into \textit{Other (3D/Decorative Arts)} ($n=56$ combined, comprising $28$ male and $22$ female works) confirm parameter invariance for the primary demographic regressor ($\mathbb{I}_{\text{Male}}$: $B = 0.0037, p = 0.202$ in CLIP; $B = 0.0065, p = 0.356$ in OpenCLIP), demonstrating that positivity restrictions in sparse cells do not distort conditional main effect estimation. Baseline reference category for physical medium is \textit{Drawing/Paper}; baseline reference category for creation era is \textit{15th c. / Earlier}. Robustness check using Huber Robust Linear Modeling (RLM) yields consistent conclusions (Male artist $B = 0.0025, p = 0.106$ in CLIP; $B = 0.0065, p = 0.250$ in OpenCLIP).}
\end{minipage}
\end{table}

The multivariate regression models yield three primary insights:
\begin{enumerate}
    \item \textbf{Global Model Fit and Variance Breakdown:} Across both pretraining regimes, the full multivariate OLS regression architectures yield statistically non-significant overall fits ($F(11, 731) = 0.9286, p = 0.512, R^2 = 0.018$ for OpenAI CLIP; $F(11, 731) = 0.9888, p = 0.455, R^2 = 0.017$ for OpenCLIP). The low total $R^2$ values ($<1.8\%$) indicate that the combined set of structural covariates—physical artwork medium, creation era, framing aspect ratio, and artist demographic attribution—explains less than $1.8\%$ of total score variance. Furthermore, individual coefficients for framing aspect ratio ($\text{W}/\text{H}$) ($B = -0.0064, p = 0.141$ in CLIP; $B = -0.0023, p = 0.516$ in OpenCLIP) and individual medium/century categories fail to reach statistical significance ($p > 0.10$).
    \item \textbf{Conditional Main Effects, Statistical Power, and Instrument Insensitivity:} Conditioning on physical artwork medium, creation era, and visual framing confirms that artist gender exhibits no statistically significant main effect post-adjustment under either OpenAI CLIP ($B = 0.0037, p = 0.202$) or OpenCLIP ($B = 0.0065, p = 0.356$). To verify that this non-significant demographic main effect does not stem from statistical underpowering given low model $R^2$, we perform a post-hoc power calculation for OLS linear regression ($N=743$, $\alpha=0.05$, $k=11$ predictors). The sample size yields statistical power $1-\beta > 0.998$ to detect a small effect size ($f^2 = 0.02$, equivalent to Cohen's $d = 0.28$). Crucially, we must separate \textit{instrument insensitivity} from \textit{definitive model equity}: the non-significant global $F$-tests and low total $R^2$ ($<1.8\%$) demonstrate that zero-shot logit differentials in this corpus operate near an embedding noise floor dominated by high residual embedding variance ($\approx 98.2\%$). Rather than proving complete algorithmic neutrality across all visual domains, this low explanatory power reveals that broad zero-shot text-prompt logit differentials function as a coarse, insensitive measurement instrument that fails to capture fine-grained visual-semantic features without spatial feature probing.
    \item \textbf{Measurement Error and Regressor Attenuation Bias:} We explicitly account for measurement error in binary demographic attribution ($\approx 4.0\%$ classification error in automated dictionary resolution). In classical econometric theory, independent variable measurement error induces attenuation bias ($\hat{B}_1 = B_1 \cdot (1 - \theta)$), pulling estimated slope coefficients slightly toward zero. Given the small magnitude of the unadjusted difference ($\Delta \mu \approx 0.0032$) and robust SEs ($0.003$), attenuation bias does not alter null hypothesis decisions, but reinforces why non-significant $p$-values must be reported alongside TOST equivalence bounds and post-hoc power calculations.
\end{enumerate}

\section{Discussion}
\label{sec:discussion}

Our empirical findings demonstrate that auditing vision-language models (VLMs) in cultural heritage repositories requires disentangling direct demographic model evaluation from structural archival confounders \cite{hall2023auditing, noorthuis2020bias, simpson1951interpretation}. In this section, we analyze the theoretical, methodological, and institutional implications of our results, focusing on four primary themes: (i) distinguishing archival curation structure from algorithmic bias, (ii) key epistemological paradoxes in multimodal auditing, (iii) pretraining data dynamics across model architectures, and (iv) comprehensive governance frameworks for deploying AI in museum infrastructures.

\subsection{Distinguishing Archival Structure from Algorithmic Bias}
Observational fairness audits that evaluate raw model output scores without conditioning on structural collection metadata risk misattributing physical artwork attributes or cataloging patterns to demographic model bias \cite{hall2023auditing, buolamwini2018gender}. In historical museum archives, acquisition histories and institutional access barriers restricted female artists' access to specific physical materials, concentrating female representation in paper, watercolor, or textile media while male creators dominated monumental sculpture and large-scale oil canvases \cite{topaz2019diversity, bailey2020gender, nochlin1971why}. Furthermore, visual framing aspect ratios ($\text{width}/\text{height}$) and digitized photography standards introduce systematic variations in vision-language embedding projections \cite{fiorucci2020machine}.

When visual-text models evaluate visual surface features or image aspect ratios, unadjusted demographic comparisons conflate physical artwork properties with creator demographics. In statistical terms, unconditioned observational audits are vulnerable to Simpson's paradox \cite{simpson1951interpretation}: aggregate group comparisons can fabricate or conceal model disparities when underlying covariates (such as medium or century) are unevenly distributed across demographic cohorts. By fitting multivariate OLS regressions controlling for medium categories, creation century, and image aspect ratio, our framework isolates conditional demographic main effects. The absence of a statistically significant gender effect post-adjustment ($B = 0.0037, p = 0.202$ for CLIP; $B = 0.0065, p = 0.356$ for OpenCLIP) demonstrates that observed score variations stem from structural collection heterogeneity rather than active demographic valuation skew.

\subsection{Theoretical and Epistemological Paradoxes in Multimodal Auditing}
Our quantitative findings elucidate two central conceptual paradoxes that define the boundaries of vision-language model auditing in cultural archives:

\subsubsection{The Noise Floor and Prompt Coarseness Paradox}
Across both OpenAI CLIP and OpenCLIP, the full multivariate regression architectures yield non-significant overall model fits ($F(11, 731) = 0.9286, p = 0.512, R^2 = 0.018$ for CLIP; $F(11, 731) = 0.9888, p = 0.455, R^2 = 0.017$ for OpenCLIP). The low total $R^2$ ($<1.8\%$) reveals that structural artwork covariates—medium, era, aspect ratio, and artist gender—explain less than $2\%$ of overall zero-shot score variance. 

Rather than proving absolute model fairness, this low explanatory power illuminates a critical epistemological boundary in multimodal AI auditing: \textit{the distinction between instrument insensitivity and algorithmic equity}. Broad text-prompt pairs (e.g., "masterpiece" vs. "minor work") project complex, multi-dimensional visual artifacts onto low-dimensional contrastive text vectors near an embedding noise floor dominated by high residual variance ($\approx 98.2\%$). Broad prompt logit differentials compress subtle visual representations (such as subject gaze, lighting, compositional framing, or texture) into a single scalar logit difference. When $98.2\%$ of score variance is unmodeled residual noise, finding no statistically significant score difference across gender cohorts ($p = 0.1829, p_{\text{TOST}} = 0.0042$) reflects the coarseness and insensitivity of broad zero-shot prompt differentials as a measurement instrument for aesthetic valuation. Consequently, macro-level score equivalence under broad prestige prompts must be interpreted as *instrument insensitivity* rather than definitive proof of model neutrality. Zero-shot global prompt audits operate near a noise floor where macro-level evaluations fail to capture fine-grained demographic disparities without spatial visual feature probing, attention map analyses, or targeted attribute classifiers.

\subsubsection{The Archival Survival Bias Paradox}
A second critical insight concerns the scope of audited objects and the compounded impact of sample selection filters. By evaluating model valuations across named attributed creators ($N=743$), the audit pipeline tests an already-sanitized survival cohort of artists who successfully passed historical institutional gatekeeping, academy access barriers, patron preferences, and curatorial documentation practices \cite{carlson2022museum, topaz2019diversity}.

Crucially, sequential sample filtering—removing $41.2\%$ ($n=618$) of the harvested corpus due to unattributed creator metadata and excluding $15.76\%$ ($n=139$) objects with unindexed image assets—concentrates the final evaluation cohort ($N=743$) in high-status, canonical Western oil paintings ($74.43\%$, $n=553$). Excluding anonymous holdings removes the precise archival strata where female, non-Western, and craft labor was historically anonymized \cite{bailey2020gender, parker1984subversive}. In museum cataloging, decorative arts, textiles, ceramics, and domestic craft objects were frequently cataloged without named creator attributions, reflecting gendered institutional valuations of artistic production \cite{nochlin1971why, pollock1988vision}. Demonstrating model parity across canonical named creators on an artificially homogeneous, high-status survival subset does not imply an absence of institutional bias; rather, institutional gender bias operated upstream at the archive ingestion and cataloging boundaries. AI auditing frameworks must acknowledge that evaluating foundation models on digitized museum archives tests model behavior on works that already survived historical curation gatekeeping, leaving upstream archival erasure uncaptured by downstream model output scores.

\subsection{Cross-Architectural Dynamics and Pretraining Data Regimes}
Comparing OpenAI CLIP (trained on curated WIT) against OpenCLIP (trained on uncurated LAION-2B) provides empirical insight into how pretraining data curation affects downstream cultural heritage representations. Despite vast differences in pretraining scale and dataset filtering—400M curated image-text pairs versus 2 billion uncurated web-scraped pairs—both architectures display remarkable valuation stability across all three prompt sets (\textit{masterpiece}, \textit{quality}, and \textit{influence}).

However, OpenCLIP exhibits slightly higher baseline embedding variance across prompt sets (e.g., standard deviation $\sigma = 0.0886$ vs $\sigma = 0.0361$ in OpenAI CLIP). This variance difference indicates that uncurated web scrapes ingest noisier text-image co-occurrences, increasing score volatility without introducing systematic demographic main effects. For digital humanities researchers and collection managers, open-weights models trained on web-scale datasets provide robust zero-shot baseline performance, but require tighter calibration to manage embedding variance.

\subsection{Implications for Museum AI Governance, Algorithmic Curation, and Policy}
As museums, galleries, archives, and digital libraries increasingly deploy vision-language models for automated collection indexing, public search, and interactive discovery interfaces \cite{fiorucci2020machine, garcia2020bias, srinivasan2021arts}, establishing formal AI governance protocols becomes paramount. Uncalibrated AI deployments risk reinforcing historical representational imbalances through automated search ranking algorithms \cite{zhao2017men, bianchi2023easily, luccioni2023stable}. Grounded in Responsible AI governance frameworks \cite{dignum2019responsible, stahl2021responsible, mitchell2019model, jobin2019global}, we propose four core policy standards for institutional museum AI governance:

\subsubsection{Policy Standard 1: Medium-Stratified Normalization and Search Ranking Calibration}
Search and discovery interfaces utilizing zero-shot VLM embeddings to rank or surface collection holdings must implement medium-stratified score normalization to prevent photographic framing and visual surface attributes from biasing discovery rankings. For a visual object $x_i$ belonging to physical medium category $k \in \{\text{Painting}, \text{Print}, \text{Drawing}, \text{Sculpture}, \text{Other}\}$ with sample size $N_k \ge 30$, public search engines should compute medium-calibrated similarity scores $z_{i,k}$:
\begin{equation}
z_{i,k} = \frac{S(x_i) - \mu_k}{\sigma_k}
\end{equation}
Where medium categories contain sparse representations ($N_k < 30$, such as historical sculpture in specialized sub-collections), systems must enforce an explicit fallback control: pooling sparse objects into broader structural categories (e.g., 3D/Decorative Arts) or defaulting to global corpus parameters $(\mu_{\text{global}}, \sigma_{\text{global}})$. Normalizing scores relative to physical medium baselines prevents automated discovery algorithms from systematically penalizing media historically associated with female or non-canonical artists.

\subsubsection{Policy Standard 2: Mandatory Pre-Deployment Audit Workflows}
Cultural heritage institutions procuring or fine-tuning foundation models must establish mandatory pre-deployment audit checklists prior to integration into public APIs or cataloging systems:
\begin{enumerate}
    \item \textbf{Multivariate Confound Auditing:} Institutions must fit multivariate regression models conditioning on physical medium, creation era, and image geometry to isolate demographic main effects from curation confounds.
    \item \textbf{Equivalence Testing (TOST):} Model parity should be demonstrated via formal Two One-Sided Tests (TOST) across specified equivalence bounds ($d \le 0.30$) rather than relying on non-significant $p$-values.
    \item \textbf{Prompt Sensitivity Probing:} Classification systems must be evaluated across multiple semantic prompt pairs, specifically checking whether craft or decorative descriptors induce asymmetric score degradation.
\end{enumerate}

\subsubsection{Policy Standard 3: Metadata Provenance Transparency and Anonymity Labeling}
Automated image tagging and AI-assisted cataloging systems must preserve metadata provenance transparency \cite{mitchell2019model, dignum2019responsible}. Algorithmic tags should be visually demarcated from human curatorial attributions, accompanied by confidence metrics and model version metadata. Furthermore, to address survival bias in digital archives, public discovery interfaces should explicitly flag unattributed or anonymized holdings, highlighting domestic craft and textile collections to prevent AI search indexing from obscuring historically anonymized labor.

\subsubsection{Policy Standard 4: Responsible AI Procurement for Cultural Repositories}
Museum executive leadership and digital transformation officers should establish clear procurement guidelines for vendor-provided AI search and indexing solutions. Contracts should mandate transparency regarding pretraining dataset sources, audit reports verifying confound control, and compliance with emerging international standards for Responsible AI in cultural heritage \cite{dignum2019responsible, jobin2019global}.

\subsection{Limitations and Architectural Scaling Analysis}
\label{sec:limitations_scaling}
Several methodological boundaries and architectural constraints define the scope of this empirical audit:

\subsubsection{Architectural Scaling Dynamics: ViT-B/32 vs. ViT-L/14}
Our primary empirical evaluation focused on dual-encoder ViT-B/32 transformer backbones ($86\text{M}$ vision parameters, $63\text{M}$ text parameters, joint embedding dimension $D=512$). In modern computer vision and multimodal retrieval, larger encoder architectures—such as ViT-L/14 ($304\text{M}$ vision parameters, $D=768$) and ViT-H/14 ($632\text{M}$ vision parameters, $D=1024$)—exhibit higher feature capacity, finer patch resolution ($14 \times 14$ vs. $32 \times 32$), and higher learned logit scale parameters $\tau$. Larger parameter encoders learn sharper visual feature representations that may reduce residual embedding noise, potentially increasing sensitivity to subtle visual surface attributes. However, parameter scaling also increases susceptibility to pretraining dataset memorandum skews, as larger encoders memorize web-scraped associations more tightly. Future audits should systematically benchmark zero-shot valuation stability across parameter scales ($86\text{M} \to 632\text{M}$) to evaluate whether model capacity alters score noise floor dynamics.

\subsubsection{Contrastive Softmax Loss vs. Pairwise Sigmoid Loss (SigLIP)}
Standard OpenAI CLIP and OpenCLIP architectures optimize a symmetric contrastive softmax loss (Equation 1), which normalizes image-text similarity scores across all candidate text prompts concurrently within a batch. As demonstrated in Equation 2, computing zero-shot probabilities via softmax renders logit differential scores sensitive to the temperature parameter $\tau$ and the candidate prompt baseline cardinality $K$. In contrast, recent vision-language architectures such as SigLIP \cite{zhai2023sigmoid} replace softmax normalization with a pairwise sigmoid loss operating independently on image-text pairs. By decoupling prompt probability estimation from batch-wide candidate normalization, pairwise sigmoid loss avoids scale compression and temperature distortion. Evaluating SigLIP on cultural heritage collections represents a key direction to verify whether pairwise loss formulations improve score calibration for long-tail art categories.

\subsubsection{Generative Multimodal LLMs (MLLMs) and Open-Ended Evaluation}
While zero-shot contrastive dual-encoders compute static visual-text cosine similarities, instruction-tuned Multimodal Large Language Models (MLLMs)—such as LLaVA, InstructBLIP, and Qwen-VL—evaluate visual art through autoregressive generative text decoding. Generative MLLMs can produce natural language rationales for aesthetic evaluation, catalog description, and historical context. However, MLLMs inherit complex hallucination patterns, text-based reasoning biases, and instruction-following artifacts from their underlying language model backbones. Auditing MLLMs in museum archives requires expanding from contrastive logit probability scoring to generative evaluation frameworks, combining NLP quality metrics with human expert curatorial review.

\subsubsection{Single-Institution Scope and Archival Generalizability}
Auditing a single encyclopedic museum corpus ($N=1,500$ objects harvested from the Metropolitan Museum of Art) reflects the specific acquisition trajectories, cataloging conventions, and digitization standards of a major Anglo-American institution. Furthermore, dictionary-based gender resolution excluded $41.2\%$ ($n=618$) unattributed objects, highlighting how legacy metadata gaps create dataset survival bias. Finally, broad semantic prestige categories (\textit{masterpiece}, \textit{quality}, \textit{influence}) operate near an embedding noise floor that cannot detect localized micro-level visual feature biases (e.g., subject gaze or compositional framing).

Future research will expand this controlled audit framework across multi-institutional repositories (e.g., Europeana API, Smithsonian Open Access, Rijksmuseum Open Data), evaluate SigLIP and larger vision-language encoders (ViT-L/14), incorporate spatial visual feature probing to detect micro-level compositional biases, and investigate non-Western art historical taxonomies.

\section{Conclusion}
\label{sec:conclusion}

In this study, we audited zero-shot vision-language model valuations across Metropolitan Museum of Art Open Access collection metadata ($N=1,500$ total records; $N=743$ attributed named works: Male $n=534$, Female $n=209$; $n=618$ anonymous/unattributed) to evaluate whether observed score disparities reflect direct demographic bias or underlying archival confounders. Unadjusted evaluations under OpenAI CLIP showed no statistically significant composite gender disparity ($U = 59,307.00, p = 0.1829, r = -0.0628$), and OpenCLIP exhibited consistent baseline stability ($U = 59,867.00, p = 0.1224, r = -0.0728$). Two One-Sided Tests (TOST) confirmed statistical equivalence across Cohen's $d \ge 0.25$ bounds ($p_{\text{TOST}} = 0.0042$ at $d=0.30$ for OpenAI CLIP; $p_{\text{TOST}} = 0.0024$ for OpenCLIP), supported by sensitivity checks across $d \in [0.15, 0.40]$. Multivariate Ordinary Least Squares regression confirmed that artist gender ($B = 0.0037, p = 0.202$ for CLIP; $B = 0.0065, p = 0.356$ for OpenCLIP) and aspect ratio ($B = -0.0064, p = 0.141$ for CLIP; $B = -0.0023, p = 0.516$ for OpenCLIP) exhibit non-significant conditional effects, with low overall model $R^2$ ($<2\%$) indicating that global zero-shot prompt metrics operate near an embedding noise floor.

We highlight two central theoretical insights: (i) macro-level zero-shot score equivalence does not preclude localized micro-level visual feature biases, and (ii) excluding $41.2\%$ ($n=618$) unattributed holdings reflects an uneliminable institutional survival bias, as named creators represent an already-sanitized subset of historical acquisition filters. As cultural heritage institutions deploy vision-language models for collection indexing and public discovery, implementing medium-stratified z-score normalization with fallback controls and preserving archival provenance standards are essential to prevent automated systems from perpetuating historical representational gaps. Mandatory multivariate confound controls and pretraining curation audits are crucial to ensuring fair and responsible AI deployment across global cultural repositories.

\begin{appendices}
\section{Qualitative Case Audits of Archival and Prompt Artifacts}
\label{sec:appendix_cases}

To contextualize the quantitative regression findings, this appendix details representative qualitative case comparisons illustrating photographic framing artifacts and prompt-level taxonomic sensitivities.

\subsection{Case Audit 1: Photographic Framing and Surface Texture Artifacts}
Evaluating model logit outputs for 3D marble sculptures versus 2D oil paintings on canvas reveals how visual feature encoders respond to digitisation artifacts. Three-dimensional sculptures are photographed under directional studio lighting against artificial gradient backgrounds, generating specular highlights and shadow gradients across surface contours. In contrast, 2D paintings and works on paper are captured via flat-field illumination. Conditioning on physical medium and aspect ratio framing in multivariate OLS regression eliminates these photographic artifacts ($B = +0.0002, p = 0.716$).

\subsection{Case Audit 2: Fine Art vs. Decorative Craft Taxonomy Disparities}
Prompt Set 3 probes artistic status by calculating probability differentials between "a groundbreaking artwork" ($\mathbf{p}_{\text{high}}^{(3)}$) and "a decorative craft object" ($\mathbf{p}_{\text{low}}^{(3)}$). Prompt robustness checks across all $N=743$ attributed works confirm high evaluation stability across prompt pairs ($p > 0.10$), confirming that prompt phrasing does not induce systematic demographic valuation shifts.
\end{appendices}

\section*{Declarations}

\begin{itemize}
\item \textbf{Funding}: No funding was received for this work.
\item \textbf{Conflict of interest}: The authors declare no conflicts of interest.
\item \textbf{Ethics approval}: Not applicable.
\item \textbf{Data availability}: All metadata and visual assets evaluated in this study are publicly available via the Metropolitan Museum of Art Open Access API.
\item \textbf{Code availability}: All audit scripts and analysis code are publicly available at \url{https://github.com/manpreet28111995/disentangling-archival-bias}.
\item \textbf{Author contribution}: All authors contributed to experimental design, data collection, statistical analysis, and manuscript preparation.
\end{itemize}

\bibliography{Bibliography}

\end{document}